\documentclass[final]{cvpr}

\usepackage{times}
\usepackage{epsfig}
\usepackage{graphicx}
\usepackage{amsmath}
\usepackage{amssymb}
\usepackage{color}
\usepackage{xifthen}
\usepackage{mathtools}
\usepackage{duckuments}
\usepackage[ruled,vlined]{algorithm2e}

\usepackage[pagebackref=true,breaklinks=true,colorlinks,bookmarks=false]{hyperref}

\def\cvprPaperID{10357} 
\def\confYear{CVPR 2021}

\newcommand{\todo}[1][]{%
\ifthenelse{\isempty{#1}}
	 {\textcolor{red}{(TODO)} \marginpar{\textcolor{red}{$\star$}}}
   {\textcolor{red}{(TODO: \marginpar{\textcolor{red}{$\star$}} #1)}}
}

\begin{document}

\title{Architectural Adversarial Robustness: The Case for Deep Pursuit}

\author{George Cazenavette\textsuperscript{1}\;\;\;\;\;Calvin Murdoch\textsuperscript{1}\;\;\;\;\;Simon Lucey\textsuperscript{1,2}\\
\textsuperscript{1}Carnegie Mellon University\;\;\;\;\;\textsuperscript{2}Argo AI\\
{\tt\small \{gcazenav,cmurdock,slucey\}@cs.cmu.edu}

}

\maketitle

\begin{abstract}
   Despite their unmatched performance, deep neural networks remain susceptible to targeted attacks by nearly imperceptible levels of adversarial noise. While the underlying cause of this sensitivity is not well understood, theoretical analyses can be simplified by reframing each layer of a feed-forward network as an approximate solution to a sparse coding problem. Iterative solutions using basis pursuit are theoretically more stable and have improved adversarial robustness. However, cascading layer-wise pursuit implementations suffer from error accumulation in deeper networks. In contrast, our new method of deep pursuit approximates the activations of all layers as a single global optimization problem, allowing us to consider deeper, real-world architectures with skip connections such as residual networks. Experimentally, our approach demonstrates improved robustness to adversarial noise.  
\end{abstract}

\section{Introduction}


Multilayer sparse approximation has 
been proposed as a
robust alternative to feed-forward neural networks \cite{romano2019adversarial}. While provably 
less sensitive to noise,
recurrent networks that implement layered basis pursuit 
accumulate independent errors 
and cannot be applied to modern large-scale architectures.
Here, we propose 
a new method of \emph{deep pursuit}, wherein all activations of the network are synchronously optimized through a global basis pursuit,
circumventing error accumulation and accounting for the skip connections commonly found in state-of-the-art network architectures.

We apply this technique to address a major weakness of deep neural networks:
despite unrivaled performance on supervised tasks,
they can be highly sensitive to certain types of data noise.
 Specifically, adversarial attacks use imperceptible targeted input perturbations to completely change a network's predictions \cite{goodfellow2015adversarial}. 
 Robustness to such attacks is mission-critical to many domains, such as security systems and autonomous vehicles.

Because 
the generalization properties of 
deep neural networks are not yet thoroughly understood, combating such adversarial attacks remains an open problem. Current state-of-the-art methods rely on specialized loss functions or training techniques. However, these methods do not explain \emph{why} some models are more susceptible to attacks than others or how to create naturally robust architectures. In this work, we 
apply
techniques from sparse approximation theory 
to design
deep neural networks
that are intrinsically more robust
to adversarial noise.
\begin{figure}
    \centering
    \includegraphics[scale=0.6]{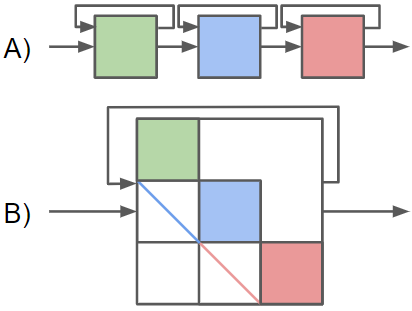}
    \caption{
    (a) Recurrent deep networks that implement layered basis pursuit have been shown to be provably more robust to adversarial noise than feed-forward alternatives. However, this method is incompatible with modern architectures. (b) We instead propose deep pursuit, which jointly infers all network activations as a single structured sparse coding problem.
    }
    \label{fig:cover}
\end{figure}

Previous works have suggested reframing each layer of a neural network as a sparse coding problem to make their outputs more robust \cite{romano2019adversarial} (Figure~\ref{fig:cover}a). However, as noted by the authors, this method accumulates error throughout the layers of the network, potentially leading to poor performance in deeper models. Additionally, this method offers no provisions for handling skip connections between layers, preventing its use for real-world network architectures.

 In order to exploit the natural robustness offered by deeper networks \cite{robustdepth} and skip connections \cite{robustskip}, we propose adapting the layer-wise pursuit algorithm introduced in \cite{romano2019adversarial} to the global view that reframes a neural network as an approximate solution to a single 
 sparse coding problem with
 block-structured 
 parameters (Figure~\ref{fig:cover}~B) \cite{murdock2020dataless}. 
We 
 optimize the outputs of all layers synchronously
 \cite{chodosh2020use},
which effectively amounts to
adding recurrent feedback connections on top of a feed-forward network. 
 By relating the entire network to a single structured sparse coding problem, our method
 does not suffer from error accumulation as the network grows deeper. Furthermore, we can even entertain residual and dense skip connections within our optimization, something not possible using layered basis pursuit. We call this method ``deep pursuit."  

Our contributions are:
\begin{enumerate}
    \item Through connections to sparse approximation theory, we illustrate how the structure of a global sparse approximation problem predicts why certain architectures are naturally more robust.
    \item Extending the method of layered basis pursuit to our global view, we propose a technique for synchronously inferring all latent activations via block coordinate descent.
    \item We show how deep pursuit outperforms layered basis pursuit by avoiding error accumulation and allowing for skip connections between layers. Experimentally, we demonstrate improved robustness to adversarial attacks on the CIFAR-10 dataset.
\end{enumerate}

\section{Related Works}
\subsection{Adversarial Examples}
In a white-box setting where an attacker has access to the model's parameters, an adversarial example can be crafted to modify a prediction by explicitly maximizing the model's loss within given bounds on the noise. Goodfellow et. al. introduced the ``fast gradient sign method'' where additive perturbations are constructed from the \emph{sign} of the gradient of the output with respect to the input \cite{goodfellow2015adversarial}.
In a purely linear model, this maximizes the change in the output. They hypothesized that this attack translates so well to neural networks because their components are all quasi-linear, despite the overall function being technically highly non-linear. 


\subsection{Adversarial Training}
Current state of the art methods for training models robust to adversarial attacks like these work by creating loss-maximizing adversarial examples at train time \cite{gan2020large, roth2019adversarial, xiong2020improved}. By doing so, the model learns to correctly classify or embed such examples. However, one caveat to this method is that it only \textit{directly} encourages robustness towards the type of attack used to generate the adversarial samples. 

Additionally, adversarial training takes much longer to converge. Since the training set is continuously being updated along with the model, the objective function is non-stationary. This results in the optimization chasing an ever-moving target, raising questions of when it is an appropriate time to stop training. Evaluating adversarial training methods also necessitates testing performance on ``seen'' versus ``unseen'' attacks. Since our proposed deep pursuit method is purely architectural, we circumvent this requirement.

\subsection{Sparse Approximation}
Sparse coding techniques are useful in signal representation tasks in that they are \emph{provably robust}. 
In contrast to feed-forward representations that may amplify input errors, iterative optimization of a sparsity-inducing objective function can be provably insensitive to input noise \cite{donoho2005stable}.
Building upon theoretical connections between feed-forward deep networks and sparse coding~\cite{papyan2017convolutional}, recent work has even shown that using a supervised sparse encoder for classification tasks theoretically bounds the adversarial error \cite{sulam2020adversarial}.

The 
robustness of sparse coding techniques relies on the redundancy of overcomplete representations. 
Their effectiveness is determined by 
the mutual coherence of the reconstruction dictionaries, or the maximum absolute normalized inner product of the atoms used in sparse linear combinations for approximating input data \cite{romano2019adversarial, sulam2020adversarial}. Small mutual coherence leads to dictionaries that are closer to orthogonal. 
Murdock and Lucey \cite{murdock2020dataless} developed a method of analyzing the global mutual coherence of deep neural networks by viewing the parameters of all layers as a single structured matrix. Adding more layers with denser connections between them decreases an architecture-dependent lower bound on the global mutual coherence. Through correlations with generalization capacity, this provided an explanation for why deeper networks and those with skip connections are more naturally robust, leading to improved generalization performance without overfitting. 
\subsection{Low-Rank Representations}
Another method of achieving robustness is by exploiting low-rank embeddings of the data \cite{awasthi2020adversarial}. It has even been shown that dropout, an implicit method of encouraging robustness with noise, is related to low-rank weight matrix factorization \cite{cavazza2018dropout}. 

\section{Architectural Robustness}

Shallow iterative sparse approximation with layered basis pursuit has been shown to be provably more robust than a feed-forward neural network layer \cite{romano2019adversarial}. 
By adapting this theory towards a global view of a deep neural network as a single structured sparse coding problem, we aim to construct neural networks that are even more robust.

\subsection{Neural Networks as Sparse Coding}

\begin{figure}
    \centering
    \includegraphics[scale=0.6]{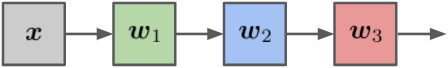}
    \caption{A feed-forward neural network can be reframed as a cascade of sparse coding problems with solutions approximated via layered thresholding pursuit.}
    \label{fig:LTH}
\end{figure}

To view a neural network as an approximate solution to a sparse coding problem, we must first see the link between proximal operators and non-linear activation functions. Because its trust region encourages sparsity in the trained parameters, the $\ell_1$ regularizer is often used as a surrogate for minimizing the (intractable) $\ell_0$ norm~\cite{mutualcoherence}. The proximal operator, a generalization of projection onto constraints, is a tool used in convex optimization to optimize objectives with non-differentiable penalty functions~\cite{parikh2014proximal}. The proximal operator of the $\ell_1$ norm with weight $\lambda > 0$ yields the elementwise soft thresholding operator:
\begin{equation}
    \phi_\lambda(\boldsymbol{x}) = 
    \begin{cases}
    \boldsymbol{x} - \lambda & \boldsymbol{x} > \lambda\\
    0 & -\lambda \leq \boldsymbol{x} \leq \lambda\\
    \boldsymbol{x} + \lambda & \boldsymbol{x} < \lambda
    \end{cases}
    \label{eq:thresholding}
\end{equation}

Papayan et. al. \cite{papyan2017convolutional} showed that if we also apply a non-negative constraint, the resulting proximal operator $\tilde{\phi}_\lambda$ in Eq.~\ref{eq:reluprox} is equivalent to the Rectified Linear Unit (ReLU), a nonlinearity commonly used in many state-of-the art deep networks, with a negative bias of $\lambda$.
\begin{equation}
    \tilde{\phi}_\lambda(\boldsymbol{x}) = \underset{\boldsymbol{w} \geq 0}{\arg\min} \tfrac{1}{2}\|\boldsymbol{x} - \boldsymbol{w}\|^2_2 + \lambda \|\boldsymbol{w}\|_1 = \mbox{ReLU}(x - \lambda)
    \label{eq:reluprox}
\end{equation}

As such, we can then reframe a single layer of a neural network as the approximate solution of the following non-negative sparse coding problem (non-negative LASSO \cite{lasso}) solved via soft-thresholding pursuit:
\begin{equation}
    \min_{\boldsymbol{w} \geq 0} \tfrac{1}{2}\|\boldsymbol{x} - \mathbf{B}\boldsymbol{w}\|^2_2 + \lambda \|\boldsymbol{w}\|_1
    \label{eq:lasso}
\end{equation}

A feed-forward chain network (without skip connections) with ReLU activations can then be interpreted as layered soft-thresholding pursuit, a cascade of approximate solutions to sparse-coding problems (Figure \ref{fig:LTH}):
\begin{equation}
    f(\boldsymbol{x}) = \tilde{\phi}_{\lambda_l}(\mathbf{B}_l^T\hdots \tilde{\phi}_{\lambda_2}(\mathbf{B}^T_2 \tilde{\phi}_{\lambda_1}(\mathbf{B}^T_1 \boldsymbol{x}))\hdots)
    \label{eq:thresholdpursuit}
\end{equation}
Here, $\boldsymbol{x}$ is an input vector and $\mathbf{B}_i$ are (dense or convolutional) parameter dictionaries.


In this setting, the parameters are learned as
\begin{equation}
    \underset{\mathbf{A},\{\mathbf{B}_{j}\}, \{\lambda_j\}}{\arg\min}\sum_{i=1}^{n}J(\mathbf{A}^Tf(\boldsymbol{x}_{i}),y_{i})
\end{equation}
where $J$ is a loss function (e.g. cross-entropy), $\mathbf{A}$ contains the parameters of a linear classifier, and $y_i$ is the target for sample $\boldsymbol{x}_i$. One drawback of this classic feed-forward design is that small perturbations in the input $\boldsymbol{x}$ can be amplified through the layers leading to large changes in the output $f(\boldsymbol{x})$
for increased sensitivity to adversarial noise.

By expressing neural networks as 
cascades of approximate sparse coding problems,
we can 
create more robust models
by reducing the noise sensitivity of each individual layer in the network.
\subsection{Local Iterations}
\begin{figure}
    \centering
    \includegraphics[scale=0.6]{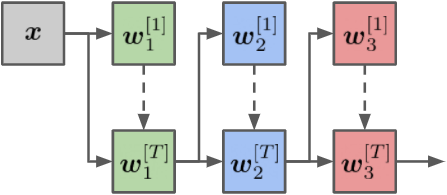}
    \caption{Instead of thresholding pursuit, Romano et. al. \cite{romano2019adversarial} proposed solving each layer's sparse coding problem with an iterative algorithm. However, this method of layered basis pursuit suffers from error accumulation and cannot account for skip connections.}
    \label{fig:layeredbasispursuit}
\end{figure}
One of the most popular algorithms used to solve the LASSO problem is the Iterative Shrinking and Thresholding Algorithm (ISTA) \cite{fista}. ISTA's iterative update takes the following form by first taking a negative (reconstruction loss) gradient step then applying the proximal operator defined in Eq. \ref{eq:thresholding}:
\begin{equation}\label{ista}
    \boldsymbol{w}^{t+1} = \phi_\lambda(\boldsymbol{w}^t - \mathbf{B}^T(\mathbf{B}\boldsymbol{w}^t - \boldsymbol{x}))
\end{equation}
When considering non-negative ISTA, we simply replace the soft thresholding operator $\phi_\lambda$ with the non-negative soft thresholding operator $\tilde{\phi}_\lambda$.

Adopting a sparse-coding view of deep learning, Romano et. al. \cite{romano2019adversarial} re-framed a deep neural network as a cascade of sparse approximation problems solved with an iterative basis pursuit algorithm.  Specifically, for an $l$-layer neural network $f$ such that $f(\boldsymbol{x})=\boldsymbol{w}_l$, for each layer $j$, we have that
\begin{equation}
    \boldsymbol{w}_j := \underset{\boldsymbol{w} \geq 0}{\arg\min}\tfrac{1}{2}\|\boldsymbol{w}_{j-1} - \mathbf{B}_j \boldsymbol{w}\|_2^2 + \lambda_j \|\boldsymbol{w}\|_1
    \label{eq:layerwise}
\end{equation}
where $\boldsymbol{w}_{j-1}$ is the output of the previous layer,
$\boldsymbol{w}_{0}=\boldsymbol{x}$, 
and $\boldsymbol{w} \geq 0$ constrains all values of $\boldsymbol{w}$ to be non-negative. Semantically, $\boldsymbol{w}$ are coefficients used to reconstruct the coefficients of the previous layer, just as ISTA optimizes coefficients to reconstruct a signal given a dictionary.

To solve this layer-wise optimization problem, we implement proximal gradient descent where the gradient of the smooth reconstruction loss, $\boldsymbol{g}_j$, is
\begin{equation}
    \boldsymbol{g}_j = \frac{\partial}{\partial \boldsymbol{w}_j}\|\boldsymbol{w}_{j-1} - \mathbf{B}_j \boldsymbol{w}_j\|_2^2 = \mathbf{B}_j^T(\mathbf{B}_j \boldsymbol{w}_j - \boldsymbol{w}_{j-1})
    \label{eq:istagrad}
\end{equation}
and the algorithm is initialized with the feed-forward soft-thresholding pursuit approximation 
\begin{equation}
    \boldsymbol{w}_j^{[0]} = \tilde{\phi}_{\lambda_j}(\mathbf{B}_j^T\boldsymbol{w}_{j-1}^{[0]}) = \mbox{ReLU}(\mathbf{B}_j^T\boldsymbol{w}_{j-1}^{[0]} - \lambda_j)
\end{equation}
which is iteratively updated by step size $\gamma_j$ as
\begin{equation}
    \boldsymbol{w}_j^{[t]} = \tilde{\phi}_{\lambda_j}(\boldsymbol{w}_{j}^{[t-1]} - \gamma_j \mathbf{B}_j^T(\mathbf{B}_j \boldsymbol{w}_j^{[t-1]} - \boldsymbol{w}_{j-1}^{[t]}))
    \label{eq:localupdate}
\end{equation}

Here, $\gamma_j$ is initialized to $\tfrac{1}{L_j}$ where $L_j$ is the conservative Lipschitz constant for guaranteed convergence rate derived in \cite{chodosh2020use}. 
However, since we do not iterate until convergence, we introduce a trainable parameter $\beta_j \in (0,1]$ that allows the network to automatically learn a larger step size such that $\gamma_j = \tfrac{1}{\beta_j L_j}$.

Our  implementation of the full algorithm is shown in Algorithm~\ref{alg:LBP}. In practice, it can be implemented as a recurrent network unrolled to a fixed number of iterations (Figure~\ref{fig:layeredbasispursuit})

It was hypothesized in \cite{romano2019adversarial} that 
explicitly solving a sparse coding problem 
makes the representation at each layer more stable, and our empirical results corroborate this theory. However, as noted in \cite{romano2019adversarial}, any error remaining is compounded through the subsequent layers of the network, making this method less effective for sufficiently deep networks. Furthermore, 
it is not clear how 
local iterations (layered basis pursuit) 
could be adapted
to network architectures with skip connections, making this method infeasible for modern architectures.

\begin{figure}
    \centering
    \includegraphics[scale=0.6]{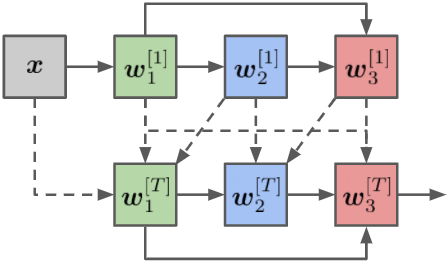}
    \caption{In deep pursuit, all layers are updated synchronously and take feedback from all adjacent layers. This method eliminates error accumulation and accounts for skip connections (as seen between layers 1 and 3).}
    \label{fig:deeppursuit}
\end{figure}

\vspace{-0.05em}
\begin{algorithm}
\label{alg:LBP}
\SetAlgoLined
 $\boldsymbol{w}_0 \leftarrow \boldsymbol{x}$\\
 \For{$j\leftarrow 1$ \KwTo $l$}{
 \For{$t  \leftarrow 1$ \KwTo $T$}{
 $\boldsymbol{w}_j^{[t]} \leftarrow \tilde{\phi}_{\lambda_j}(\boldsymbol{w}_{j}^{[t-1]} - \frac{1}{L_j \beta_j} \boldsymbol{g}_j^{[t]})$
 }
 }
 
 \caption{Inference with layered thresholding pursuit \cite{romano2019adversarial}. See Eq.~\ref{eq:istagrad} for a definition of $\boldsymbol{g}$.}
\end{algorithm}

\subsection{Deep Pursuit}

As shown in Eq. \ref{eq:thresholdpursuit}, a neural network can be expressed as a cascade of multiple sparse coding problem. Alternatively, we can view the activations of an entire network as an approximate solution to a single sparse coding problem. In this case, we can take our global loss as the sum of each layer's loss and infer all coefficients synchronously. 


By treating the parameters of the neural network as a single structured matrix, we can re-frame the global non-negative LASSO objective as 
\begin{equation}
    \underset{\{\boldsymbol{w}_j \geq 0\}}{\arg\min}\tfrac{1}{2}\sum_{j=1}^l \|\boldsymbol{w}_{j-1}-\mathbf{B}_j\boldsymbol{w}_j\|_2^2 + \lambda_j\|\boldsymbol{w}_j\|_1
    \label{eq:deeppursuit}
\end{equation}
Viewed as a structured matrix \cite{murdock2020dataless}, this can be equivalently expressed as
\begin{multline}
    \underset{\{\boldsymbol{w}_j \geq 0\}}{\arg\min} \tfrac{1}{2} \left\lVert \left[\begin{array}{c}
\boldsymbol{x}\\
\mathbf{0}\\
\vdots\\
\mathbf{0}
\end{array}\right]{-}\left[\begin{array}{cccc}
\mathbf{B}_{1} &  &  & \mathbf{0}\\
-\mathbf{I} & \mathbf{B}_{2}\\
 & \ddots & \ddots\\
\mathbf{0} &  & -\mathbf{I} & \mathbf{B}_{l}
\end{array}\right]\left[\begin{array}{c}
\boldsymbol{w}_{1}\\
\boldsymbol{w}_{2}\\
\vdots\\
\boldsymbol{w}_{l}
\end{array}\right]\right\rVert _{2}^{2}
\\
+\sum_{j=1}^{l}\lambda_j \|\boldsymbol{w}_j\|
\label{eq:globaldict}
\end{multline}
Now, instead of relying on a cascading composition of solutions, we can infer all coefficients jointly by solving a single shallow sparse coding problem. 

Structure-agnostic algorithms for shallow solutions (like soft thresholding pursuit) will not work in this case since there is no way for information to propagate from the input $\boldsymbol{x}$ to the output $f(\boldsymbol{x})$. Instead, we can now group the variables and apply the block coordinate descent algorithm introduced in \cite{blockcoord}
and applied in \cite{chodosh2020use}. This optimization is similar to that of local iterations, except that it incorporates feedback from connected layers, and every layer is updated in sequence during each iteration.

In this setting, the update for each layer $j$ during global iteration $t$ is given as
\begin{equation} 
    \boldsymbol{w}_j^{[t]} = \tilde{\phi}_{\lambda_j}(\hat{\boldsymbol{w}}_j^{[t-1]} - \tfrac{1}{\beta_jL_j}\hat{\boldsymbol{g}}_{j-1}^{[t]})
\end{equation}
where $\hat{\boldsymbol{w}}$ is an extrapolation of the previous iteration's result:
\begin{equation}
    \hat{\boldsymbol{w}}_j^{[t-1]} = \alpha_j (\boldsymbol{w}_j^{[t-1]}-\boldsymbol{w}_j^{[t-2]})
\end{equation}

The gradient of the global reconstruction error with respect to block $j$ is defined as
\begin{equation}
    \hat{\boldsymbol{g}}_{j}^{[t-1]}=\begin{cases}
\begin{gathered}\mathbf{B}_{j}^{T}(\mathbf{B}_{j}\hat{\boldsymbol{w}}_{i}^{[t-1]}-\boldsymbol{w}_{j-1}^{[t]})\quad\quad\\
\quad+(\hat{\boldsymbol{w}}_{j}^{[t-1]}-\mathbf{B}_{j+1}\boldsymbol{w}_{j+1}^{[t-1]})
\end{gathered}
 & j<l\\
\mathbf{B}_{j}^{T}(\mathbf{B}_{j}\hat{\boldsymbol{w}}_{j}^{[t-1]}-\boldsymbol{w}_{j-1}^{[t]})\quad & j=l
\end{cases}
    \label{eq:global_gradient}
\end{equation}

We see that the gradient for the global update shown in Eq. \ref{eq:global_gradient} is similar for the gradient in the layer-wise update shown in Eq. \ref{eq:localupdate}. The key distinction is that the global update takes into account feedback from the subsequent layer's result from the previous iteration because there are now two terms that include $\boldsymbol{w}_j$ since the following layer is trying to reconstruct it.

Since all parameters are updated synchronously, we no longer have the problem of error accumulation present in layered basis pursuit \cite{romano2019adversarial}. This allows us to entertain deeper networks without fear of exploding error in cascading solutions. While the robustness of layered basis pursuit is determined by the mutual coherence of each layer \cite{romano2019adversarial}, the robustness of global deep pursuit is instead determined by the mutual coherence of the global structured dictionary.

We give an outline of our deep pursuit algorithm in Algorithm \ref{alg:deeppursuit}. Note that the inner and outer loops are reversed so that the output of each layer is now updated at each iteration. In contrast to the layered pursuit algorithm in Figure~\ref{fig:layeredbasispursuit}, observe that the deep pursuit algorithm in Figure~\ref{fig:deeppursuit} allows for skip connections. The feedback connections between layers alleviate the information bottleneck between layers, facilitating more effective backpropogation in deeper networks and inducing a less coherent global dictionary.
\begin{algorithm}
\label{alg:deeppursuit}
\SetAlgoLined
 $\boldsymbol{w}_0 \leftarrow \boldsymbol{x}$\\
 \For{$j\leftarrow 1$ \KwTo $l$}{
 $\boldsymbol{w}_j^{[0]} \leftarrow \tilde{\phi}_{\lambda_j}(\mathbf{B}_j^T\boldsymbol{w}_{j-1}^{[0]}) $
 }
 \For{$t \leftarrow 1$ \KwTo $T$}{
    \For{$j \leftarrow 1$ \KwTo $l$}{
        $\boldsymbol{w}_j^{[t]} \leftarrow \tilde{\phi}_{\lambda_j}(\hat{\boldsymbol{w}}_j^{[t-1]} - \tfrac{1}{\beta_jL_j}\hat{\boldsymbol{g}}_{j-1}^{[t]})$
    }
 }
 \caption{Inference with deep pursuit. See Eqs.~\ref{eq:global_gradient} and \ref{eq:gradskip} for definitions of $\hat{\boldsymbol{g}}$.}
\end{algorithm}

\subsection{Global Iterations with Skip Connections}

Skip connections between layers have become key components of nearly all state-of-the-art architectures~\cite{resnet, densenet}. From the perspective of sparse approximation, denser skip connections between layers induce global dictionary structures with lower mutual coherence \cite{murdock2020dataless}, leading to improved robustness even with feed-forward approximations.
While the method of layered basis pursuit \cite{romano2019adversarial} gives no clear way to incorporate these skip connections, deep pursuit can be naturally adapted to support skip connections between layers. 

To accomplish this, we can modify the global
structured dictionary matrix
in Eq.~\ref{eq:globaldict}
to account for the general skip connections by including additional off-diagonal blocks of parameters:
\begin{equation}
\mathbf{B} =
\begin{bmatrix}
\mathbf{B}_{1} &  &  & \mathbf{0}\\
-\mathbf{B}_{21}^{T} & \mathbf{B}_{2}\\
\vdots & \ddots & \ddots\\
-\mathbf{B}_{l1}^{T} & \cdots & -\mathbf{B}_{l(l-1)}^{T} & \mathbf{B}_{l}
\label{eq:blockskip}
\end{bmatrix}
\end{equation}
Taking advantage of the block lower-diagonal structure, approximate feed-forward inference for sparse coding problems with dictionaries like these can proceed incrementally through the network~\cite{murdock2020dataless}. Specifically, the output $\boldsymbol{w}_j$ of layer $j$ can be found given all previous outputs as:
\begin{align}
\boldsymbol{w}_{j} & = \tilde{\phi}_{\lambda_j}\Big(\mathbf{B}_{j}^{T}\sum_{k=1}^{j-1}\mathbf{B}_{jk}^{T}\boldsymbol{w}_{k}\Big)
\label{eq:skip_feedforward}
\\
 & \approx\underset{\boldsymbol{w}_{j}}{\arg\min}\Big\Vert \mathbf{B}_{j}\boldsymbol{w}_{j}-\sum_{k=1}^{j-1}\mathbf{B}_{jk}^{\mathsf{T}}\boldsymbol{w}_{k}\Big\Vert _{2}^{2}+\lambda_j \|\boldsymbol{w}_j\|_{1}\nonumber
\end{align}    

This equation gives the general case for dense connections between every layer. 
For example, a residual connection between layers $n$ and $m$ is represented by $\mathbf{B}_{mn}=\mathbf{I}$ since the coefficients of layer $n$  are simply added to the pre-activations of layer $m$. Further details on these structured dictionary matrices can be found in \cite{murdock2020dataless}.

With this denser global dictionary, our gradients now include feedback from all connected layers (Figure \ref{fig:deeppursuit}):
\begin{equation}
\begin{split}
    \boldsymbol{\hat{g}}_j &= \mathbf{B}_j^T(\mathbf{B}_j\boldsymbol{w}_j - \sum_{k=1}^{j-1}\mathbf{B}^T_{jk}\boldsymbol{w}_k)\\
    &+ \sum_{j'=j+1}^l \mathbf{B}_{j'j}(\sum_{k'=1}^{j'-1}\mathbf{B}^T_{j'k'}\boldsymbol{w}_{k'}-\mathbf{B}_{j'}\boldsymbol{w}_{j'})
    \label{eq:gradskip}
\end{split}
\end{equation}

In addition to the well-known benefits of skip connections to generalization and trainability, the addition of more feedback connections between layers in our gradients permits information to propagate throughout the network faster, allowing for improved adversarial robustness with fewer iterations.
\begin{figure*}
    \centering
    \includegraphics[scale=1.0]{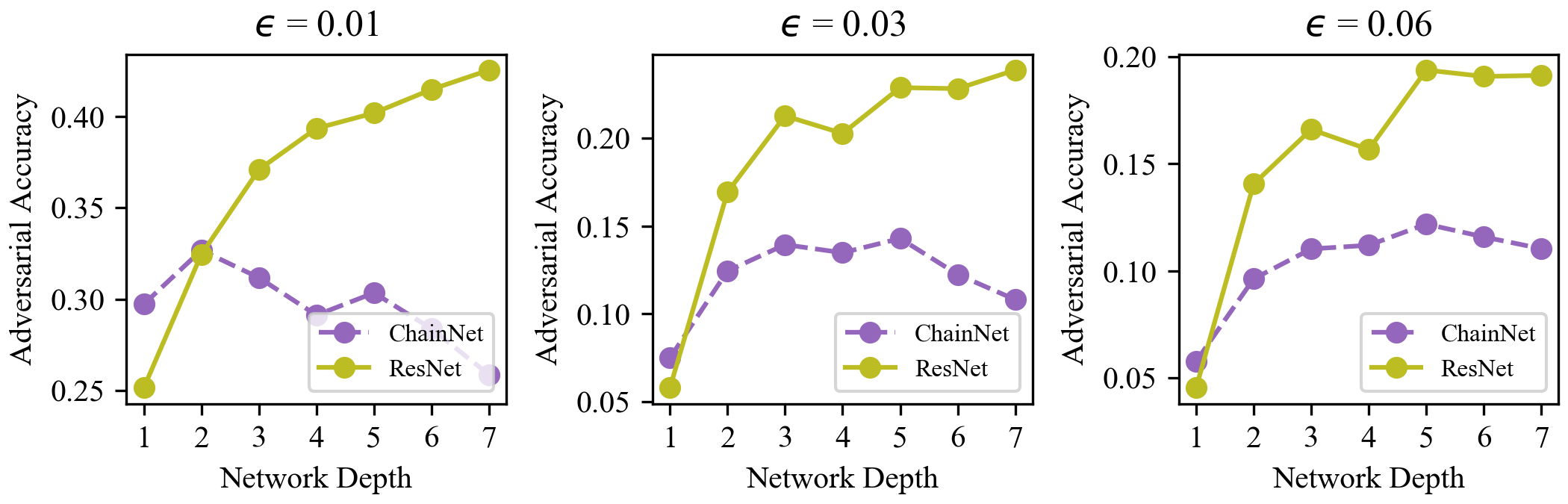}
    \caption{In the feed-forward case, deeper networks with skip connections are, in general, naturally more adversarially robust than shallow networks and those without skip connections. We propose the deep pursuit method to exploit this natural robustness since layered basis pursuit cannot.}
    \label{fig:welch}
\end{figure*}
\section{Methods and Results}
Our results highlight the improved adversarial robustness of deep pursuit over layered basis pursuit and provide theoretical insights through analysis of several sparse coding metrics, including frame potential, mutual coherence, and reconstruction error. 

For the following experiments, we train networks on the CIFAR-10 dataset \cite{cifar} and use the fast gradient sign method  \cite{goodfellow2015adversarial} with appropriate values of $\epsilon$ to generate our adversarial noise. 
Adversarial examples are constructed as
\begin{equation}
    \tilde{\boldsymbol{x}} = \boldsymbol{x} + \epsilon\cdot \mbox{sign}(\frac{\partial J(\boldsymbol{x},y)}{\partial \boldsymbol{x}})
\end{equation}
by moving each pixel in direction that would cause the loss to increase the most. We constrain our adversarial perturbations to the $\ell_\infty$ ball of radius $\epsilon$.

We use the modified ResNet architectures from \cite{murdock2020dataless} adapted to smaller input resolutions. (Further details can be found in the supplementary appendix.) These are compared against chain networks with residual connections removed, which have the same number of total learned parameters. We also use batch normalization~\cite{ioffe2015batch} adapted for our recurrent architectures by fixing the scale and offset from the feed-forward initialization for all subsequent iterations. Note that the bias corresponds to the weight of the $\ell_1$ penalty $\lambda$ in Eq.~\ref{eq:deeppursuit}, so we constrain it to be non-negative to ensure convexity. 

In our figures, ``L-TP'' refers to layered thresholding pursuit (feed-forward),  ``L-BP'' refers to layered basis pursuit (local iterations), ``DP'' refers to deep pursuit (global iterations), and ``DP-res'' refers to deep pursuit with residual connections.

\subsection{Depth, Skip Connections, and Robustness}

From the perspective of sparse approximation, network architectures 
that can induce global dictionary structures with lower mutual coherence--limited by the Welch bound--are less sensitive to input perturbations~\cite{murdock2020dataless}. Two simple ways of decreasing the Welch bound of an architecture are by making the network deeper or adding skip connections. Figure~\ref{fig:welch} shows that skip connections improve the adversarial robustness of sufficiently deep feed-forward networks. Furthermore, the residual networks become more robust as we add more layers. These two results motivate developing a new pursuit algorithm that can entertain deeper networks with residual connections.

\subsection{Recurrence by Iterative Optimization}

\begin{figure}
    \centering
    \includegraphics{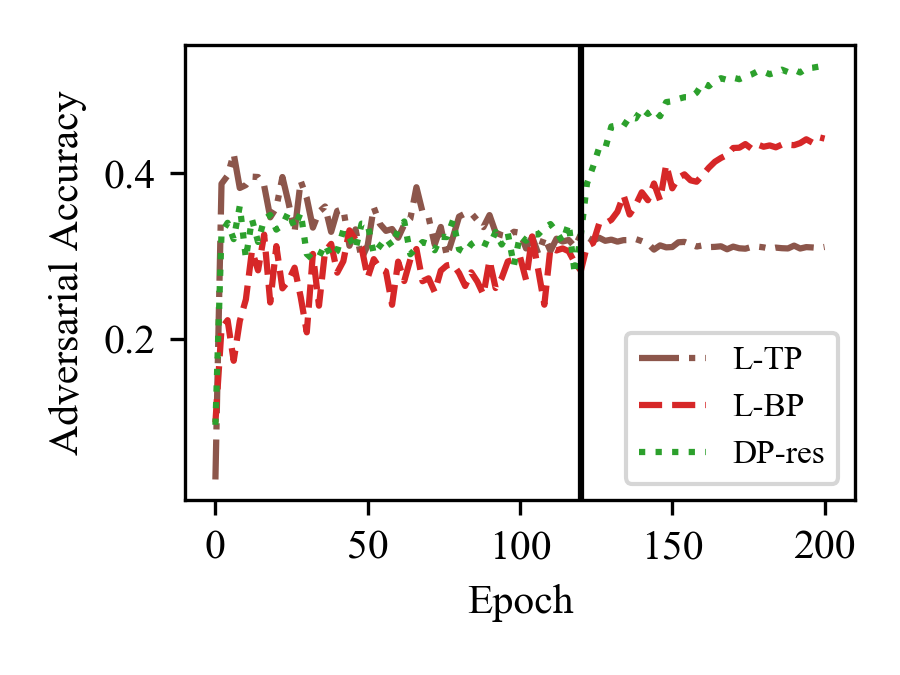}
    \caption{Most of the improvement in adversarial robustness in both layered basis pursuit (L-BP) and deep pursuit (DP) comes after the training error has converged (black vertical line). In the feed-forward case (L-TP), adversarial robustness degrades as training continues.}
    \label{fig:trainingcurve}
\end{figure}

Figure~\ref{fig:trainingcurve} shows a sample training curve from a depth-1 residual network with 10 iterations. In all our deep pursuit experiments, we found that most of the improvement in adversarial accuracy came right around when the training accuracy converged. More interestingly, the adversarial accuracy continues to improve well after both training and validation error have converged.

\begin{figure*}
    \centering
    \includegraphics[scale=1.0]{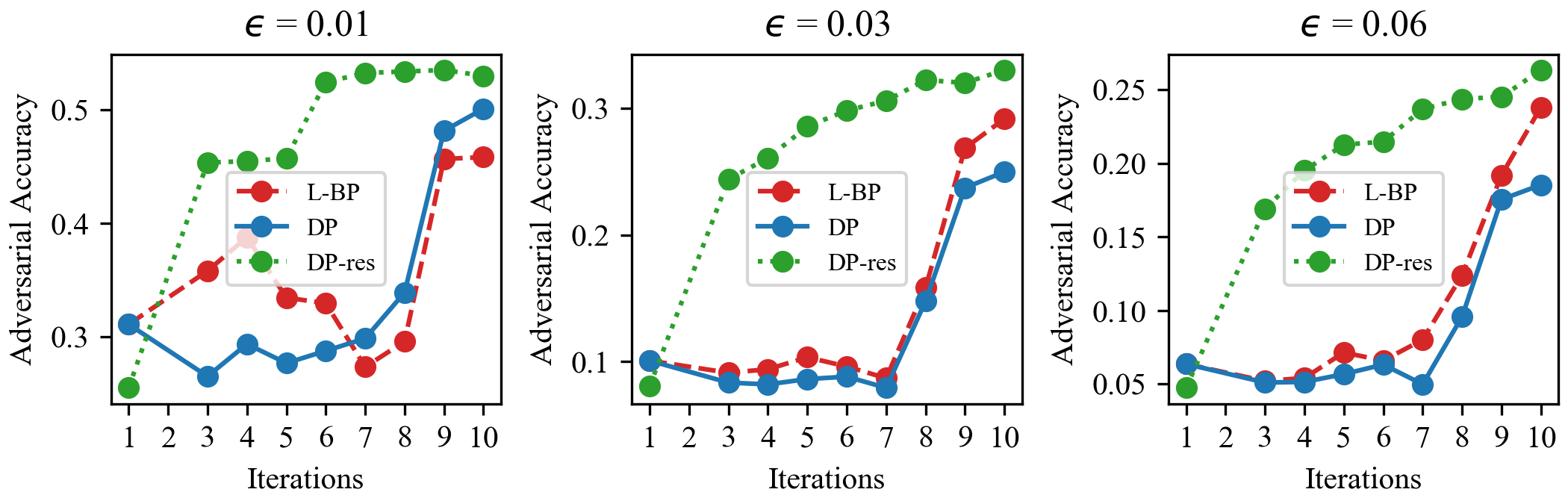}
    \caption{Deep pursuit with residual connections (DP-res) starts to outperform layered basis pursuit (L-BP) almost immediately. Without skip connections, deep pursuit (DP) requires more iterations to propagate information through the layers, so it takes longer to see noticeable improvement. (results from a depth-1 network)}
    \label{fig:iterations}
\end{figure*}

Before moving on to deeper networks with skip connections, we will establish a baseline showing that deep pursuit (global iterations) performs just as well as layered thresholding pursuit (local iterations). Observe in Figure~\ref{fig:iterations} that once we use enough global iterations, we achieve performance comparable to local iterations. We hypothesize that deep pursuit (without skip connections) requires a baseline number of iterations to achieve the full benefit of adversarial robustness because it takes an iteration for information to propagate through a layer before moving to the next.
\subsection{Effect of Residual Connections}
As neural networks become deeper, skip connections become necessary to achieve state-of-the-art performance \cite{resnet}. While local layered pursuit could not account for these skip connections, our global deep pursuit incorporates them into the optimization. Figure~\ref{fig:iterations} highlights the benefits of skip connections on adversarial robustness in iterative networks.


In addition to subverting the problem of error accumulation present in layered thresholding pursuit \cite{romano2019adversarial}, deep pursuit's compatibility with skip connections are possibly our largest contribution. Not only can we entertain larger, real-world networks, but the addition of skip connections actually makes our algorithm achieve better performance with fewer iterations than on a traditional chain network. We We hypothesize that this is because the skip connections allow information to propagate to more layers per iteration, speeding up the internal optimization.
\begin{figure}
    \centering
    \includegraphics{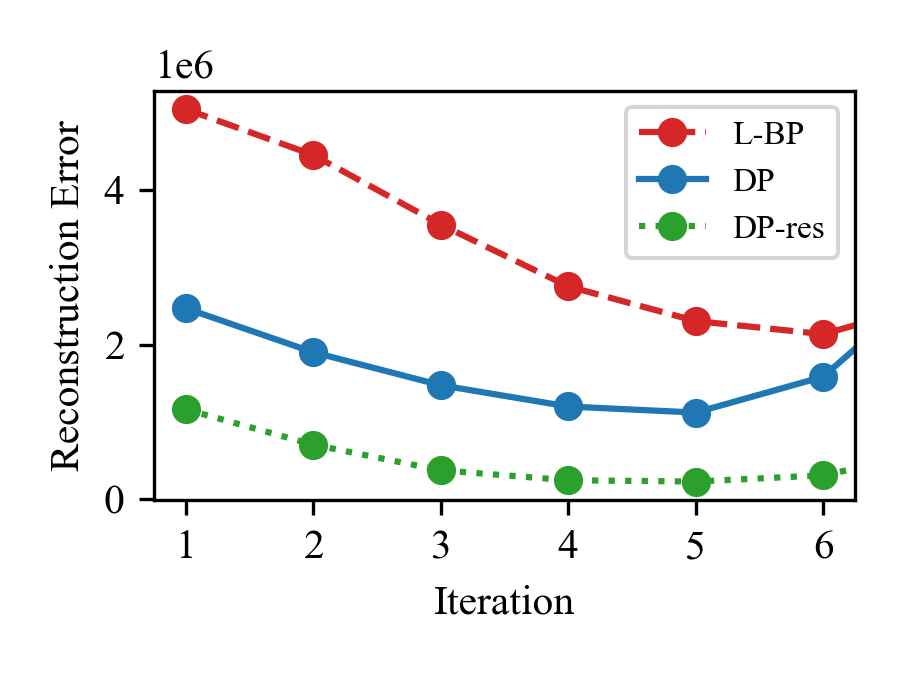}
    \caption{If we analyze the average reconstruction error of each layer per iteration (of a 10 iteration model), we observe that deep pursuit (with and without skip connections) induces more robust embeddings.}
    \label{fig:reconstruction}
\end{figure}
\subsection{Reconstruction Error}
As the main objective of sparse approximation algorithms is reconstructing a signal, the reconstruction error of our algorithms should give insight as to why one outperforms another. In our experiments, deep pursuit had, on average, lower reconstruction error per iteration than layered basis pursuit (Figure \ref{fig:reconstruction}). Adding skip connections to deep pursuit further reduced the reconstruction error per iteration.

We also saw that for all three models, the reconstruction error increased over the last few iterations, regardless of the total number. If we only used the conservative Lipschitz constant \cite{chodosh2020use} as the internal step-size for our pursuit algorithms, we would see reconstruction error strictly decrease over the iterations. However, our step size is learned based on the loss of the classification task, explaining the increase in reconstruction error over the last few iterations.

\subsection{Deep Welch Bound}
\begin{figure}
    \centering
    \includegraphics[scale=1.0]{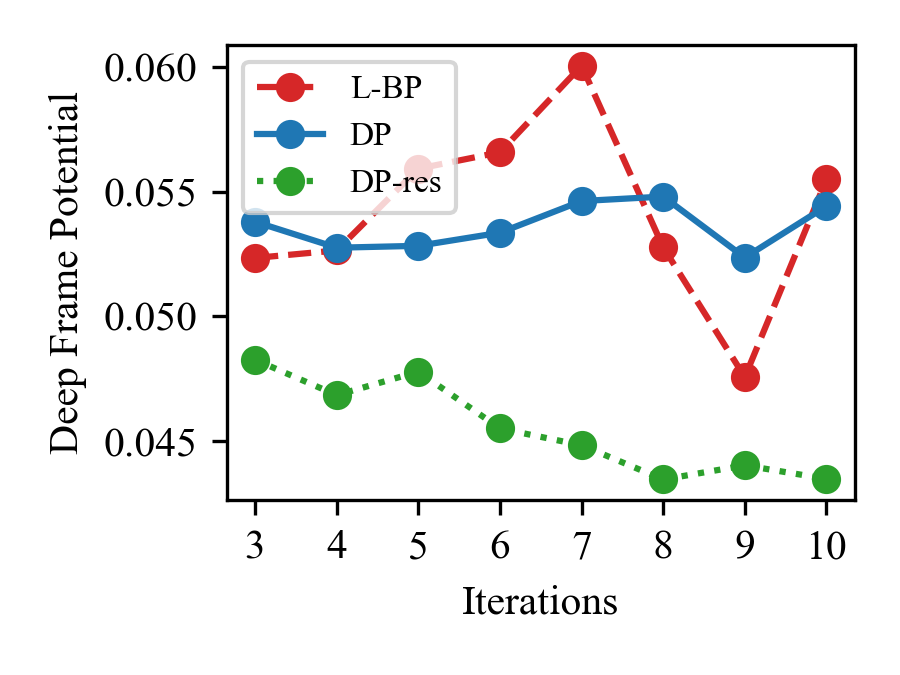}
    \caption{Deep frame potential as described in \cite{murdock2020dataless}. As we add more iterations, the deep frame potential of deep pursuit model (with residual connections) decreases, indicating a more robust solution.}
    \label{fig:deepframe}
\end{figure}
\begin{figure}
    \centering
    \includegraphics[scale=1.0]{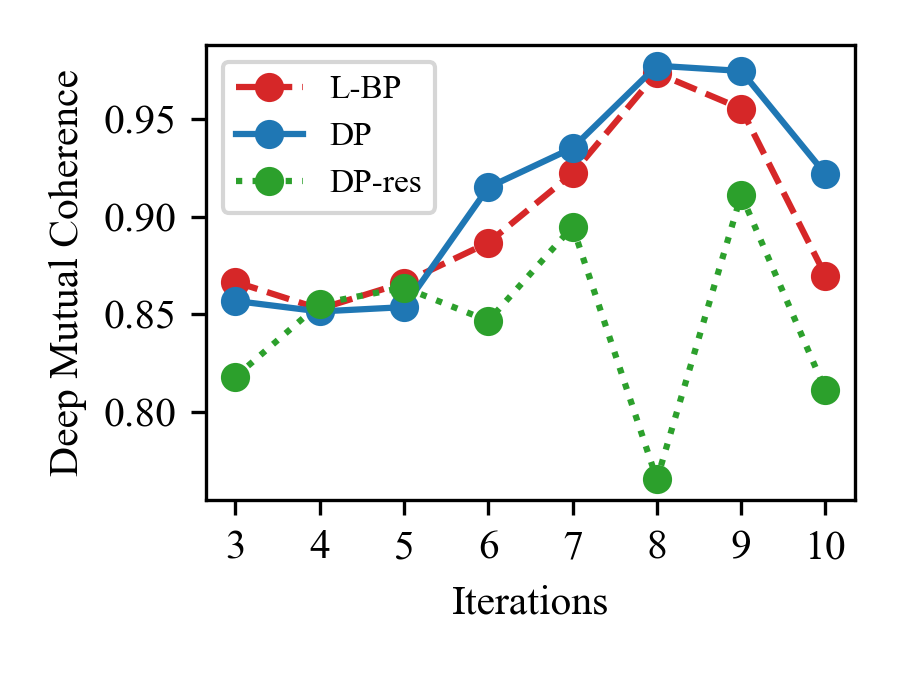}
    \caption{Deep mutual coherence as described in \cite{murdock2020dataless}. While a less stable metric than frame potential, we still see that the addition of skip connections (DP-res) induces a lower mutual coherence in our model.}
    \label{fig:deepcoherence}
\end{figure}As described in the related works, mutual coherence (maximum absolute inner product of columns) and frame potential (mean absolute inner product of columns) are measures of the sensitivity of a linear system. Layered basis pursuit and global deep pursuit (without skip connections) have the same global dictionary structure and, therefore, the same Welch bound. However, since deep pursuit optimizes the whole global dictionary synchronously, the deep frame potential is typically lower than that of layered basis pursuit~(Figure~\ref{fig:deepframe}).  Furthermore, we know that skip connections lower the Welch bound, allowing global deep pursuit to have an even smaller deep frame potential and mutual coherence. By comparing Figures~\ref{fig:deepframe} and \ref{fig:deepcoherence} with Figure~\ref{fig:iterations} we see that the deep frame potential and mutual coherence are good predictors of the adversarial robustness of a network.
\section{Conclusions}
In this work, we extend provably robust sparse approximation techniques to deep neural networks to make them more resistant to adversarial attacks. Previous work reframed each network as a separate sparse coding problem \cite{romano2019adversarial}, but this technique of layered basis pursuit is prone to error accumulation and cannot entertain modern architectures with skip connections. In contrast, our new method of deep pursuit treats the entire network as a single sparse coding problem to be solved synchronously. As such, our method avoids the issue of error accumulation. Furthermore, by viewing all network parameters as a single structured matrix, our method is easily extended to account for skip connections. Deep pursuit with skip connections consistently out-performs layered basis pursuit and induces a lower deep mutual coherence, which is theoretically tied to adversarial robustness.

Overall, we introduce a strictly architectural method of inducing adversarial robustness in modern neural networks and provide theoretical reasoning for its effectiveness. A theoretical understanding of robust architectures is a significant step towards solving the ever-changing problem of adversarial attacks. As more of modern society continues to rely on computer vision and machine learning, it is critical that we ensure the safety and robustness of these techniques.
\section{Acknowledgements}
This material is based upon work supported by the National Science Foundation Graduate Research Fellowship under Grant No. DGE1745016.
{\small
\bibliographystyle{ieee_fullname}
\bibliography{cvpr}

\begin{thebibliography}{10}\itemsep=-1pt

\bibitem{awasthi2020adversarial}
Pranjal Awasthi, Himanshu Jain, Ankit~Singh Rawat, and Aravindan
  Vijayaraghavan.
\newblock Adversarial robustness via robust low rank representations.
\newblock {\em Advances in Neural Information Processing Systems}, 33, 2020.

\bibitem{fista}
Amir Beck and Marc Teboulle.
\newblock A fast iterative shrinkage-thresholding algorithm for linear inverse
  problems.
\newblock {\em SIAM journal on imaging sciences}, 2(1):183--202, 2009.

\bibitem{cavazza2018dropout}
Jacopo Cavazza, Pietro Morerio, Benjamin Haeffele, Connor Lane, Vittorio
  Murino, and Rene Vidal.
\newblock Dropout as a low-rank regularizer for matrix factorization.
\newblock In {\em International Conference on Artificial Intelligence and
  Statistics}, pages 435--444. PMLR, 2018.

\bibitem{chodosh2020use}
Nathaniel Chodosh and Simon Lucey.
\newblock When to use convolutional neural networks for inverse problems.
\newblock In {\em Proceedings of the IEEE/CVF Conference on Computer Vision and
  Pattern Recognition}, pages 8226--8235, 2020.

\bibitem{mutualcoherence}
David~L. Donoho and Michael Elad.
\newblock Optimally sparse representation in general (nonorthogonal)
  dictionaries via l1 minimization.
\newblock {\em Proceedings of the National Academy of Sciences},
  100(5):2197--2202, 2003.

\bibitem{donoho2005stable}
David~L. Donoho, Michael Elad, and Vladimir~N. Temlyakov.
\newblock Stable recovery of sparse overcomplete representations in the
  presence of noise.
\newblock {\em IEEE Transactions on Information Theory}, 52(1), 2005.

\bibitem{gan2020large}
Zhe Gan, Yen-Chun Chen, Linjie Li, Chen Zhu, Yu Cheng, and Jingjing Liu.
\newblock Large-scale adversarial training for vision-and-language
  representation learning.
\newblock {\em arXiv preprint arXiv:2006.06195}, 2020.

\bibitem{goodfellow2015adversarial}
Ian Goodfellow, Jonathon Shlens, and Christian Szegedy.
\newblock Explaining and harnessing adversarial examples.
\newblock In {\em International Conference on Learning Representations}, 2015.

\bibitem{robustskip}
Minghao Guo, Yuzhe Yang, Rui Xu, Ziwei Liu, and Dahua Lin.
\newblock When nas meets robustness: In search of robust architectures against
  adversarial attacks.
\newblock In {\em Proceedings of the IEEE/CVF Conference on Computer Vision and
  Pattern Recognition}, pages 631--640, 2020.

\bibitem{resnet}
Kaiming He, Xiangyu Zhang, Shaoqing Ren, and Jian Sun.
\newblock Deep residual learning for image recognition.
\newblock In {\em Proceedings of the IEEE conference on computer vision and
  pattern recognition}, pages 770--778, 2016.

\bibitem{densenet}
Gao Huang, Zhuang Liu, Laurens Van Der~Maaten, and Kilian~Q Weinberger.
\newblock Densely connected convolutional networks.
\newblock In {\em Proceedings of the IEEE conference on computer vision and
  pattern recognition}, pages 4700--4708, 2017.

\bibitem{ioffe2015batch}
Sergey Ioffe and Christian Szegedy.
\newblock Batch normalization: Accelerating deep network training by reducing
  internal covariate shift.
\newblock {\em arXiv preprint arXiv:1502.03167}, 2015.

\bibitem{cifar}
Alex Krizhevsky.
\newblock Learning multiple layers of features from tiny images.
\newblock Technical report, 2009.

\bibitem{murdock2020dataless}
Calvin Murdock and Simon Lucey.
\newblock Dataless model selection with the deep frame potential.
\newblock In {\em Proceedings of the IEEE/CVF Conference on Computer Vision and
  Pattern Recognition}, pages 11257--11265, 2020.

\bibitem{papyan2017convolutional}
Vardan Papyan, Yaniv Romano, and Michael Elad.
\newblock Convolutional neural networks analyzed via convolutional sparse
  coding.
\newblock {\em The Journal of Machine Learning Research}, 18(1):2887--2938,
  2017.

\bibitem{parikh2014proximal}
Neal Parikh and Stephen Boyd.
\newblock Proximal algorithms.
\newblock {\em Foundations and Trends in optimization}, 1(3):127--239, 2014.

\bibitem{romano2019adversarial}
Yaniv Romano, Aviad Aberdam, Jeremias Sulam, and Michael Elad.
\newblock Adversarial noise attacks of deep learning architectures: Stability
  analysis via sparse-modeled signals.
\newblock {\em Journal of Mathematical Imaging and Vision}, pages 1--15, 2019.

\bibitem{roth2019adversarial}
Kevin Roth, Yannic Kilcher, and Thomas Hofmann.
\newblock Adversarial training is a form of data-dependent operator norm
  regularization.
\newblock {\em arXiv preprint arXiv:1906.01527}, 2019.

\bibitem{robustdepth}
Dong Su, Huan Zhang, Hongge Chen, Jinfeng Yi, Pin-Yu Chen, and Yupeng Gao.
\newblock Is robustness the cost of accuracy?--a comprehensive study on the
  robustness of 18 deep image classification models.
\newblock In {\em Proceedings of the European Conference on Computer Vision
  (ECCV)}, pages 631--648, 2018.

\bibitem{sulam2020adversarial}
Jeremias Sulam, Ramchandran Muthukumar, and Raman Arora.
\newblock Adversarial robustness of supervised sparse coding.
\newblock {\em Advances in Neural Information Processing Systems}, 33, 2020.

\bibitem{lasso}
Robert Tibshirani.
\newblock Regression shrinkage and selection via the lasso.
\newblock {\em Journal of the Royal Statistical Society: Series B
  (Methodological)}, 58(1):267--288, 1996.

\bibitem{xiong2020improved}
Yuanhao Xiong and Cho-Jui Hsieh.
\newblock Improved adversarial training via learned optimizer.
\newblock {\em arXiv preprint arXiv:2004.12227}, 2020.

\bibitem{blockcoord}
Yangyang Xu and Wotao Yin.
\newblock A block coordinate descent method for regularized multiconvex
  optimization with applications to nonnegative tensor factorization and
  completion.
\newblock {\em SIAM Journal on imaging sciences}, 6(3):1758--1789, 2013.

\end{thebibliography}
}
\clearpage
\section{Appendix}
\subsection{Network Architectures}
All of our networks consist of 3 ``blocks'' followed by a global average pooling layer and  a linear classifier. For a  network of depth $n$, each block consists of $n$ ``units'' composed of 2 convolutional layers. In our residual networks, there is a residual connection between each unit's input and second convolutional layer. Note that these residual connections do not increase the total number of parameters such that chain networks and residual networks of the same depth and width have the same total number of parameters.

For a network of width $m$, the convolutional layers of blocks 1, 2, and 3 have $m$, $2m$, and $4m$ filters of size $3\times 3$ respectively. After global average pooling, the linear classifier takes input of dimension 128 for width-8 networks and 64 for width-4. 
\begin{figure}
    \centering
    \includegraphics[scale=1.0]{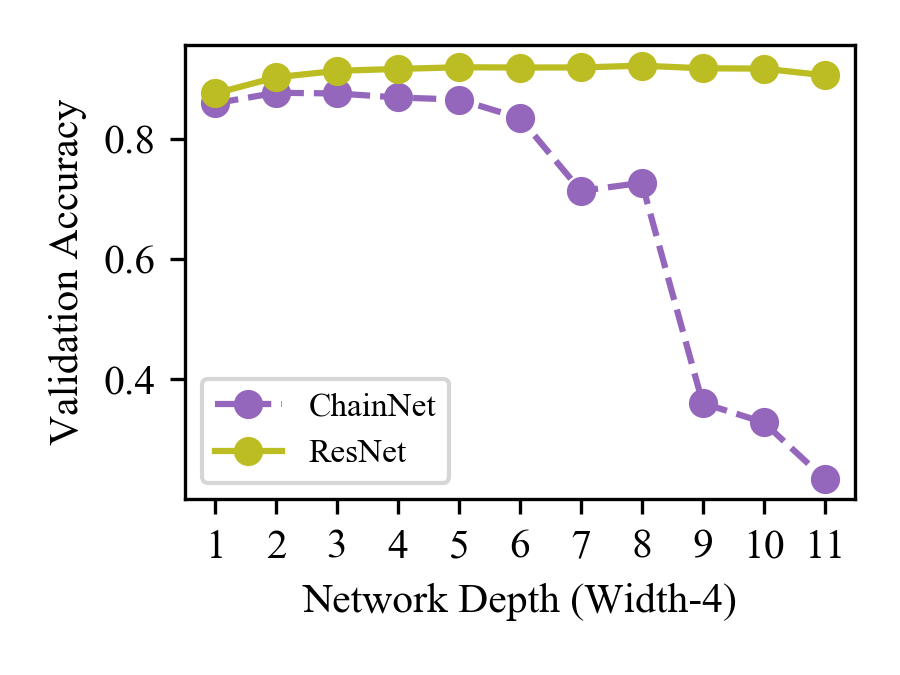}
    \caption{As we increase the network depth, we see that chain networks struggle to train. This is a  known issue with chain networks, and further motivates a new pursuit method that allows for skip connections.}
    \label{fig:welch_valid_4}
\end{figure}
\begin{figure}
    \centering
    \includegraphics[scale=1.0]{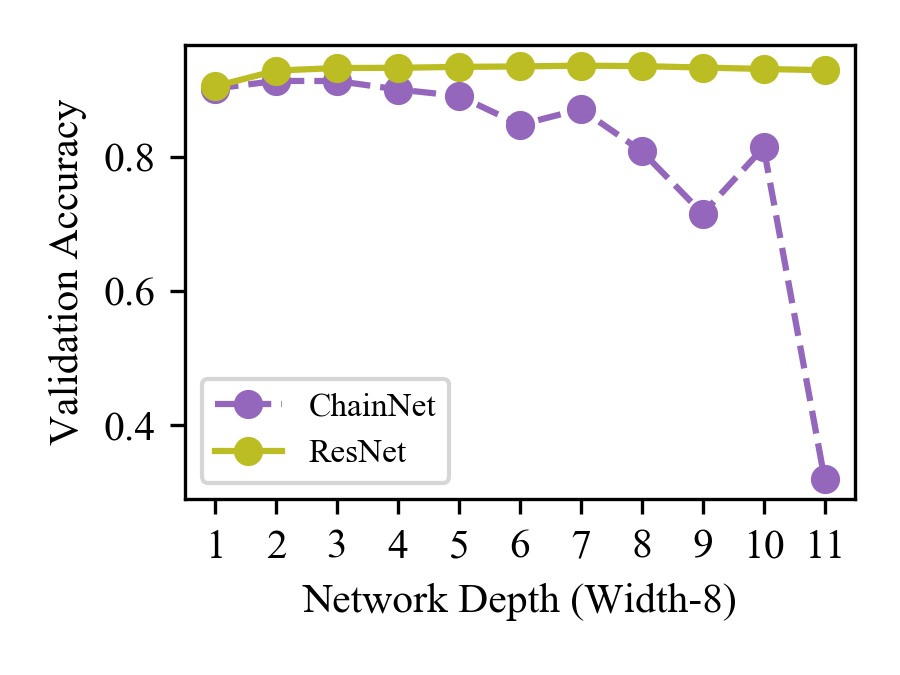}
    \caption{Likewise, the width-8 chain networks also struggle to train when they become sufficiently deep.}
    \label{fig:welch_valid_8}
\end{figure}
\subsection{Further Depth Analysis}
\begin{figure*}
    \centering
    \includegraphics[scale=1.0]{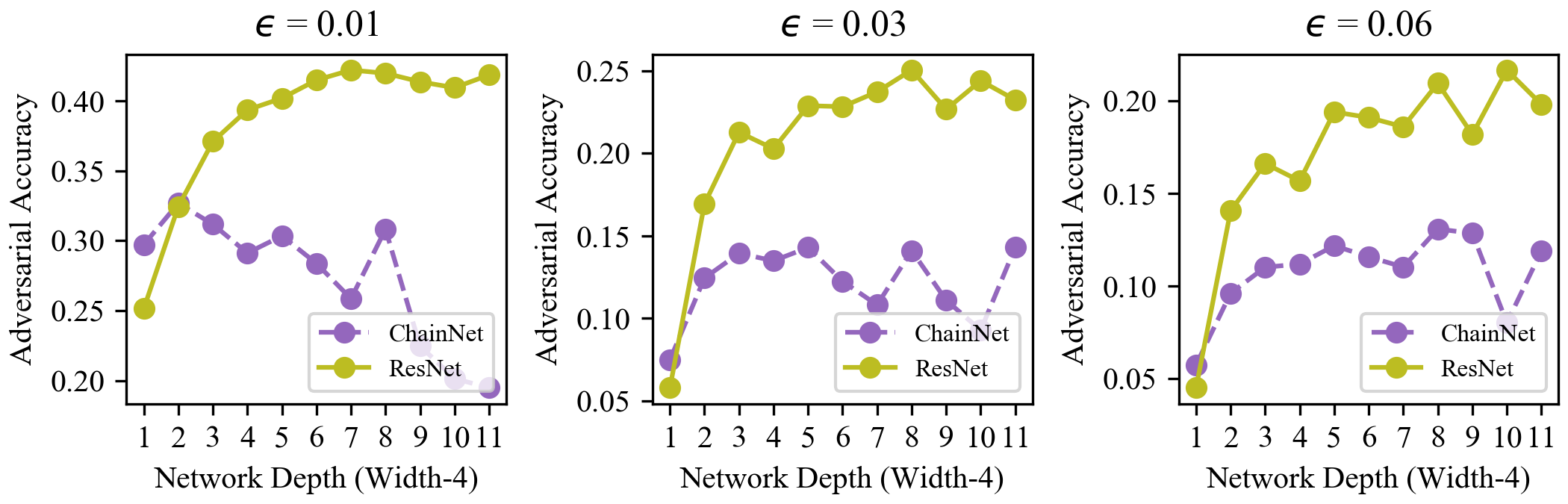}
    \caption{As observed in the body of the paper, residual networks of width-4 are consistently more adversarially robust than their chain network counterparts.}
    \label{fig:welch_4}
\end{figure*}

\begin{figure*}
    \centering
    \includegraphics[scale=1.0]{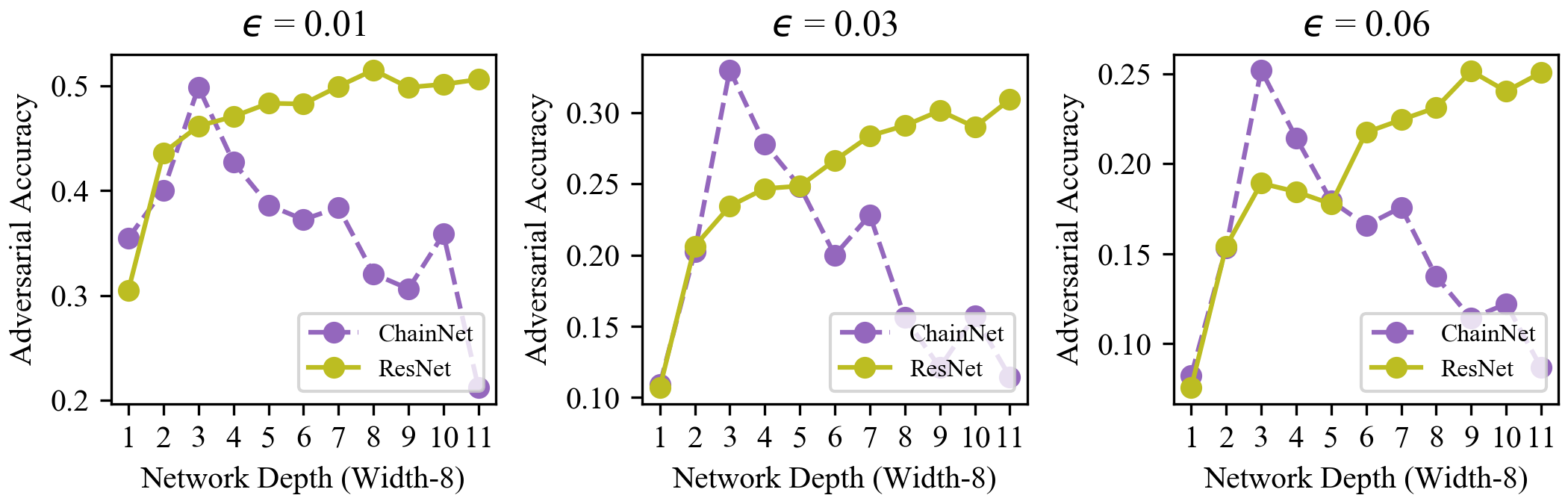}
    \caption{Here, we observe some anomalies in the depth 3 and 4 networks where the chain networks are more robust than the residual networks. We observed that these residual networks converged to zero training loss extremely quickly, resulting in no parameter updates during the part of the training process where we normally saw the biggest increase in adversarial robustnesss.}
    \label{fig:welch_8}
\end{figure*}

Figure \ref{fig:welch} shows results for feed-forward networks of width-4. Here, we include results on width-4 networks of more depths (Figures \ref{fig:welch_4}, \ref{fig:welch_valid_4}) as well as width-8 networks (Figures \ref{fig:welch_8}, \ref{fig:welch_valid_8}). The rest of the results in the body of the paper are on width-8 networks.
\end{document}